\documentclass[preprint,authoryear,a4paper]{elsarticle}
\usepackage{geometry}

\usepackage{color,soul}

\usepackage{url}
\usepackage[hidelinks]{hyperref}
\hypersetup{
    colorlinks   = true,
    urlcolor     = blue,
    linkcolor    = blue,
    citecolor    = blue
}

\emergencystretch=\maxdimen
\usepackage{graphicx}
\graphicspath{{./images/}}
\DeclareGraphicsExtensions{.pdf,.png,.jpg}
\usepackage{placeins}
\usepackage{subcaption}
\usepackage{float}
\usepackage{caption}

\usepackage{booktabs}
\usepackage[table,xcdraw]{xcolor}
\usepackage{graphicx}
\usepackage{adjustbox}
\usepackage{multirow}
\usepackage{tabularx}

\usepackage{amssymb}
\usepackage{gensymb}
\usepackage[mathscr]{euscript}
\usepackage{bm}
\usepackage{mathtools}
\usepackage{nccmath}
\usepackage{amsmath}
\usepackage{braket}
\usepackage[ruled]{algorithm2e}
\usepackage{algpseudocode}

\usepackage{lineno,hyperref}

\usepackage{lipsum}

\newcommand{\red}[1]{#1}

\journal{ISPRS Journal of Photogrammetry and Remote Sensing}

\begin{document}

\begin{frontmatter}

    \title{Depth-Enhanced Feature Pyramid Network for Occlusion-Aware Verification of Buildings from Oblique Images}

    \author[addr1]{Qing Zhu}
    \author[addr1]{Shengzhi Huang}
    \author[addr1]{Han Hu\corref{cor}}
    \cortext[cor]{Corresponding Author: han.hu@swjtu.edu.cn}
    \author[addr2]{Haifeng Li}
    \author[addr1]{Min Chen}
    \author[addr3]{Ruofei Zhong}
    \address[addr1]{Faculty of Geosciences and Environmental Engineering, Southwest Jiaotong University, Chengdu, China}
    \address[addr2]{School of Geosciences and Info-Physics, Central South University, Changsha, China}
	\address[addr3]{Beijing Advanced Innovation Center for Imaging Technology, Capital Normal University, Beijing, China}
    \begin{abstract}
        Detecting the changes of buildings in urban environments is essential.
        Existing methods that use only nadir images suffer from severe problems of ambiguous features and occlusions between buildings and other regions.
        Furthermore, buildings in urban environments vary significantly in scale, which leads to performance issues when using single-scale features.
        To solve these issues, this paper proposes a fused feature pyramid network, which utilizes both color and depth data for the 3D verification of existing buildings 2D footprints from oblique images.
        First, the color data of oblique images are enriched with the depth information rendered from 3D mesh models.
        Second, multiscale features are fused in the feature pyramid network to convolve both the color and depth data.
        Finally, multi-view information from both the nadir and oblique images is used in a robust voting procedure to label changes in existing buildings.
        Experimental evaluations using both the ISPRS benchmark datasets and Shenzhen datasets reveal that the proposed method outperforms the ResNet and EfficientNet networks by 5\% and 2\%, respectively, in terms of recall rate and precision.
        We demonstrate that the proposed method can successfully detect all changed buildings; therefore, only those marked as changed need to be manually checked during the pipeline updating procedure; this significantly reduces the manual quality control requirements.
        Moreover, ablation studies indicate that using depth data, feature pyramid modules, and multi-view voting strategies can lead to clear and progressive improvements.
        
    \end{abstract}
\end{frontmatter}


\section{Introduction}
\label{s:intro}

Buildings form the skeleton of urban environments; however, they are subject to continuous change, especially in areas where rapid urbanization takes place \citep{nyaruhuma2012verification_2d}.
As an important source of spatial infrastructure data, continuous urban management requires up-to-date building information.
It is difficult---if not impossible---to rebuild the entire dataset for each time interval; thus, it is essential to identify the changed regions. 
The existing buildings must be verified to remove demolished ones or update modified ones in the existing datasets.

Most existing building-verification methods only consider the 2D outlines of buildings, commonly using remote-sensing images to extract roof top information \citep{hussain2013change,hong2019learnable}.
For example, the normalized difference vegetation index was used to extract the roof areas \citep{rottensteiner2007building,singh2012building}, which is inevitably disturbed by high-rise vegetation near buildings; feature engineering or learned representations are also common approaches to identify roof areas \citep{sidike2016automatic,sofina2016building}, which is hard to overcome confusing textures in urban environment using only the nadir views, such as pavements, bridge decks and streets.

\red{In fact, false-alarm detection is probably the most challenging issue for change detection of buildings \citep{javed2020objectbased}.
In order to improve the robustness, some seminal works \citep{nyaruhuma2012verification_2d,xiao2015building} have been done on building verification using penta-view aerial oblique images \citep{remondino2015oblique}.}
Because of their multi-angle capabilities, both roof tops and fa\c{c}ades were considered.
The additional information provided by the fa\c{c}ade images has facilitated the detection of regular, structured features of buildings; however, some issues remain unaddressed.

\begin{figure}[H]
    \centering
    \subcaptionbox{Nadir views}[0.3\linewidth]{
        \includegraphics[width=\linewidth]{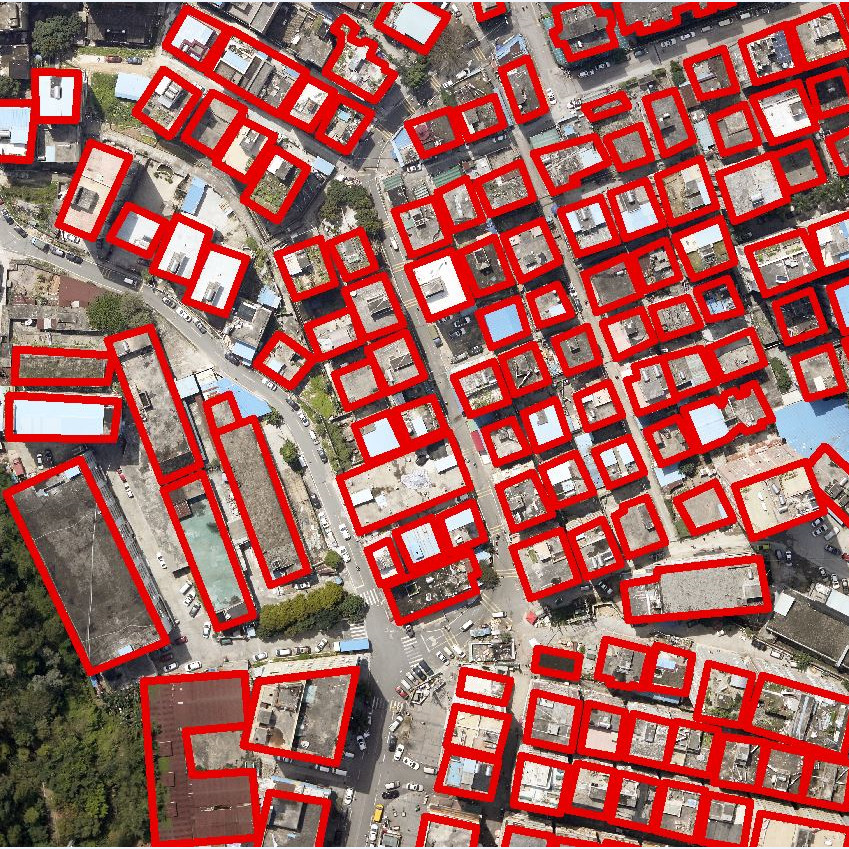}\vspace{0.3em}
        \includegraphics[width=\linewidth]{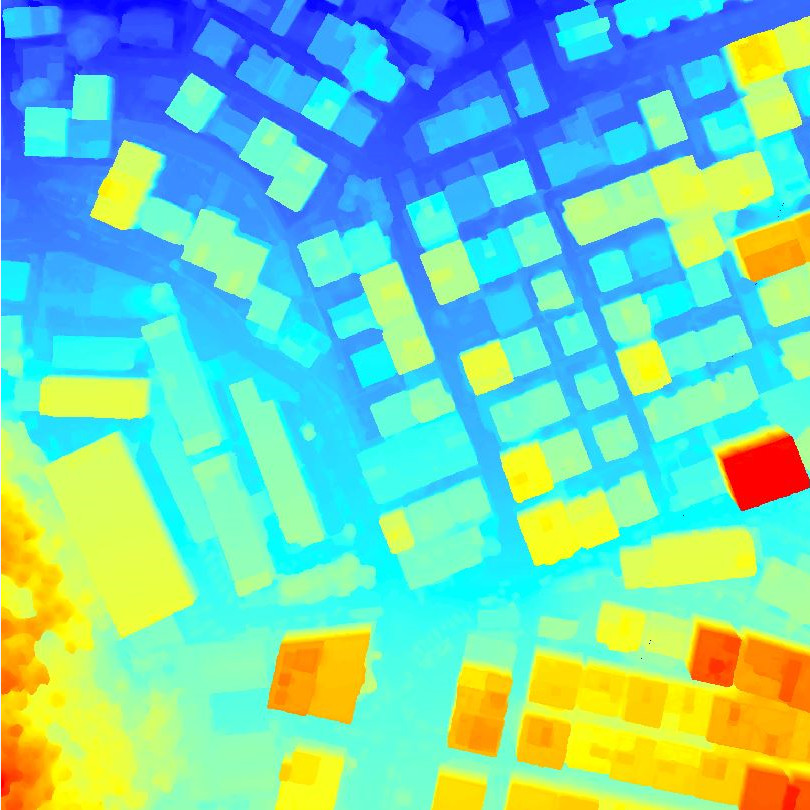}
    }
    \subcaptionbox{Ambiguous ground areas}[0.3\linewidth]{
        \includegraphics[width=\linewidth]{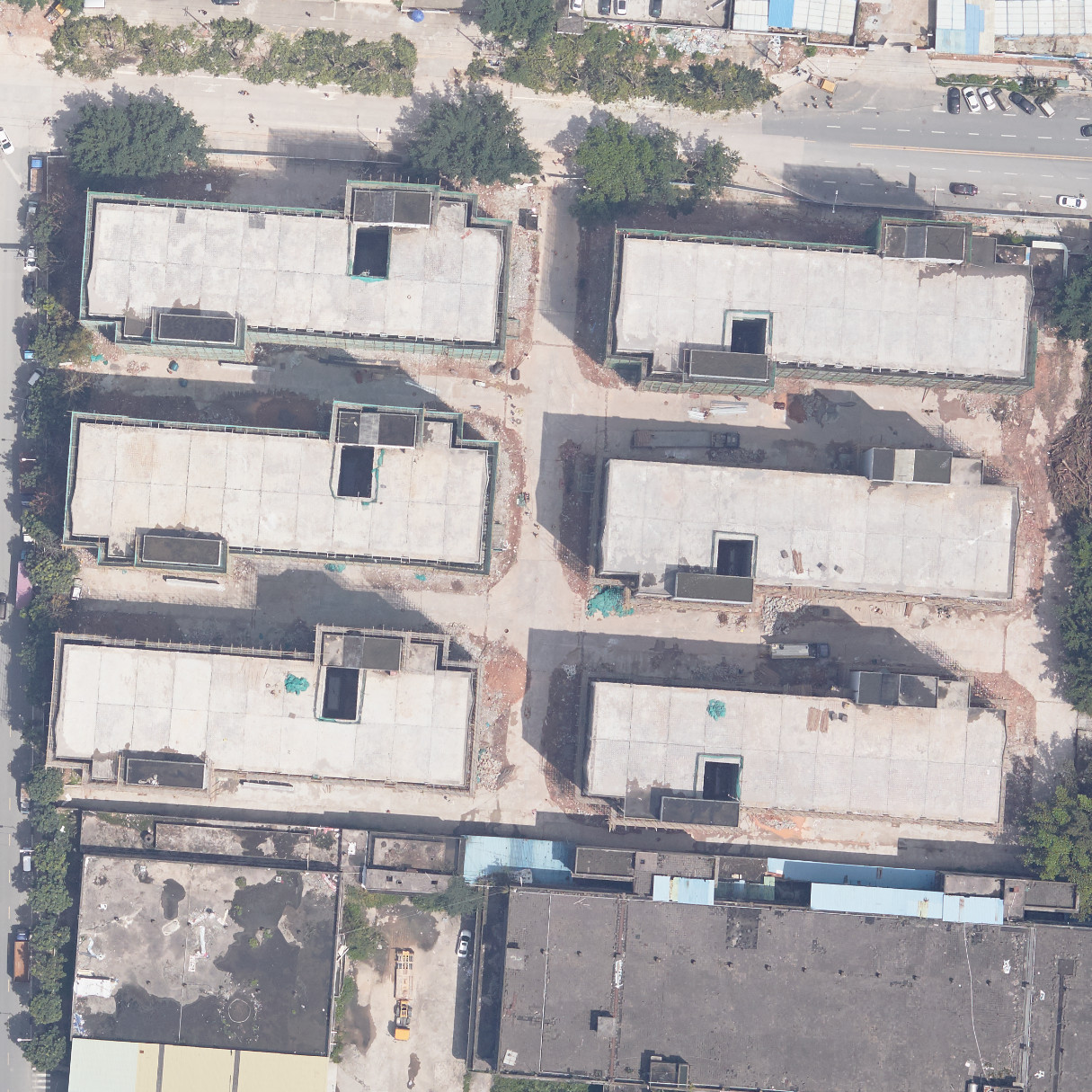}\vspace{0.3em}
        \includegraphics[width=\linewidth]{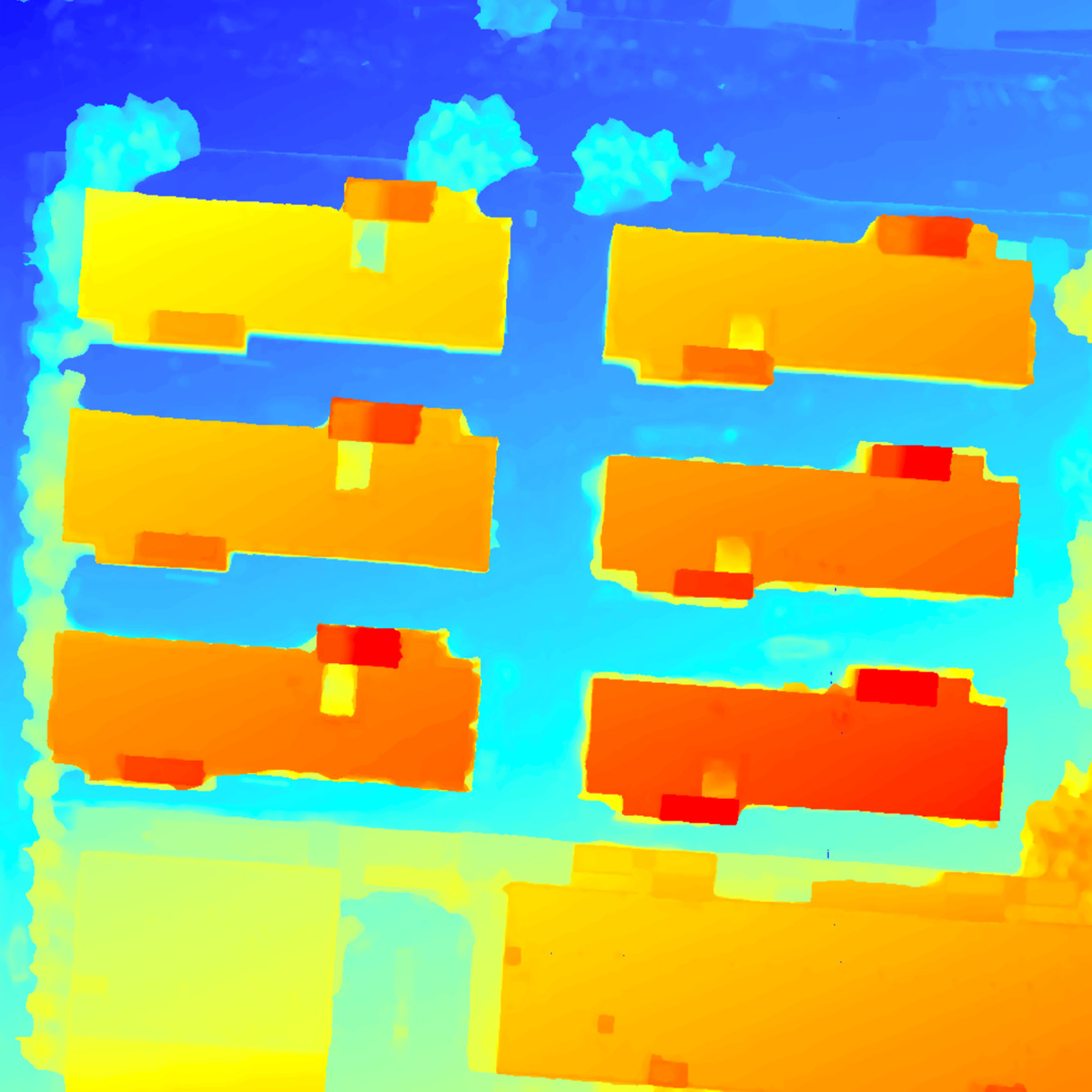}
    }
    \subcaptionbox{Ambiguous vegetation areas}[0.3\linewidth]{
        \includegraphics[width=\linewidth]{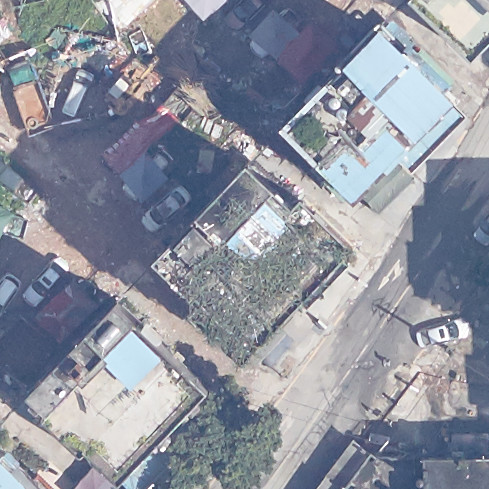}\vspace{0.3em}
        \includegraphics[width=\linewidth]{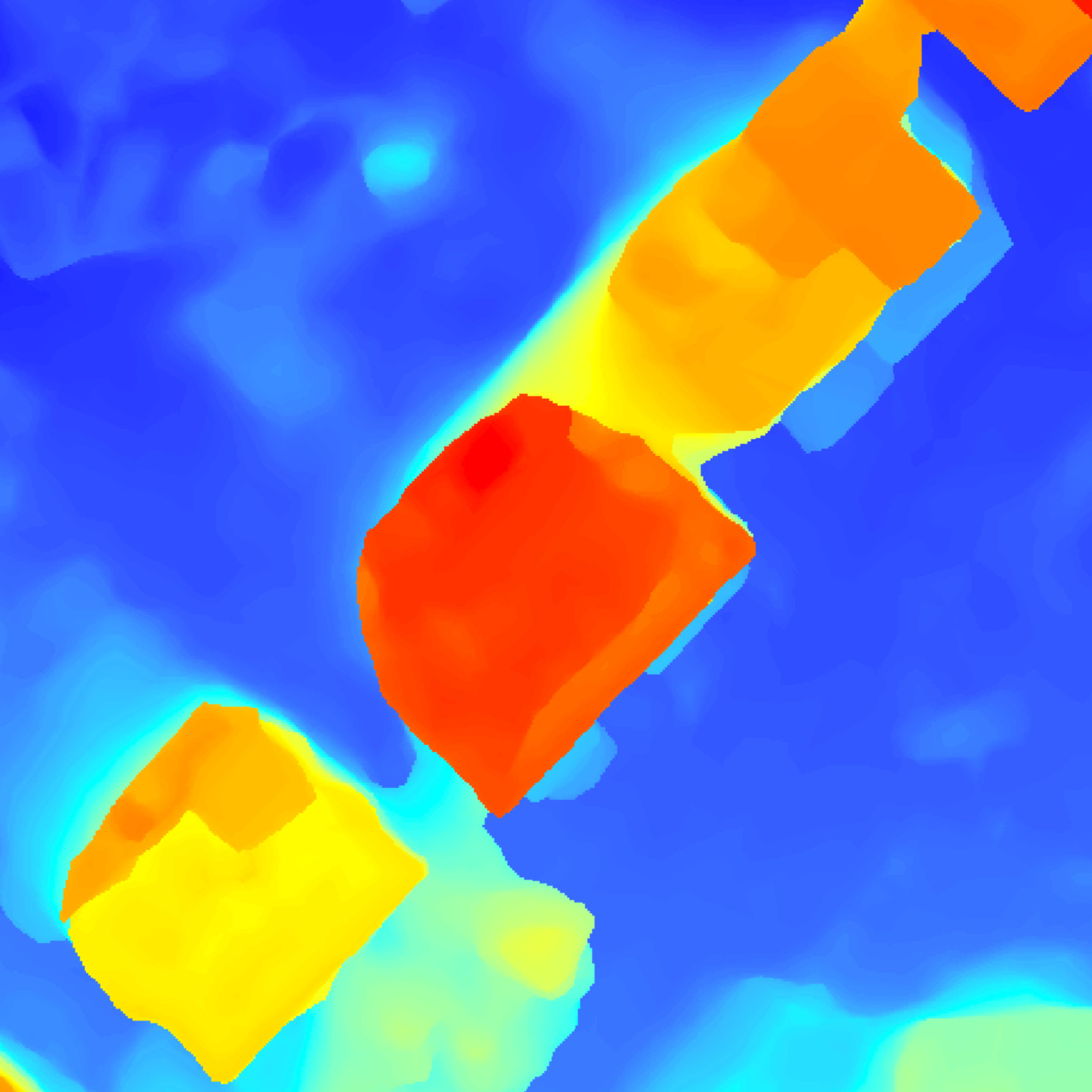}
    }
    \caption{Problems of ambiguous textures for roof areas (e.g. ground and vegetation). The first and second rows represent the color images and corresponding depth information, respectively. It should be noted that the depth information renders the building regions more salient.}
    \label{fig:problem_nadir}
\end{figure}

\begin{figure}[H]
    \centering
    \subcaptionbox{Oblique views}[0.3\linewidth]{
        \includegraphics[width=\linewidth]{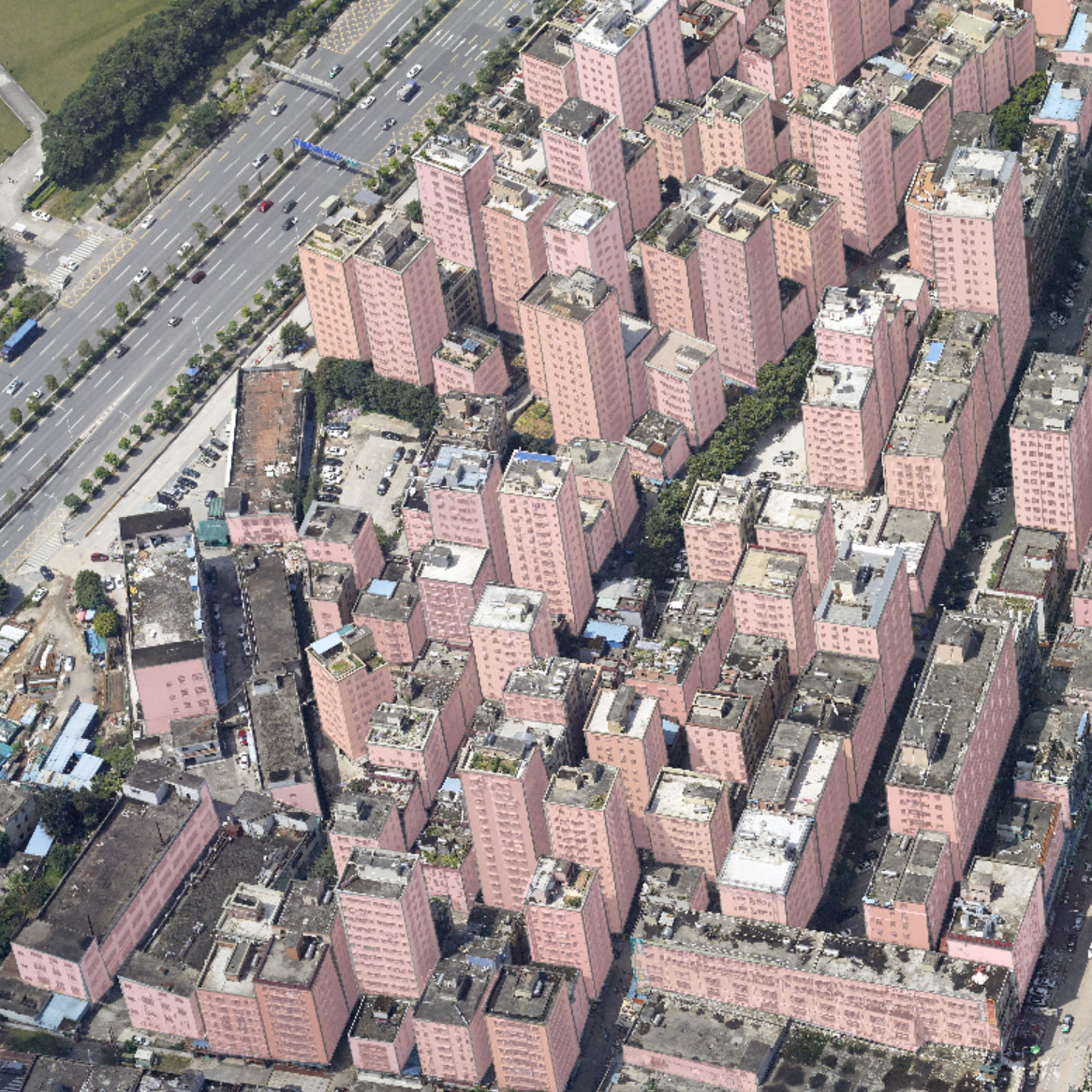}\vspace{0.3em}
        \includegraphics[width=\linewidth]{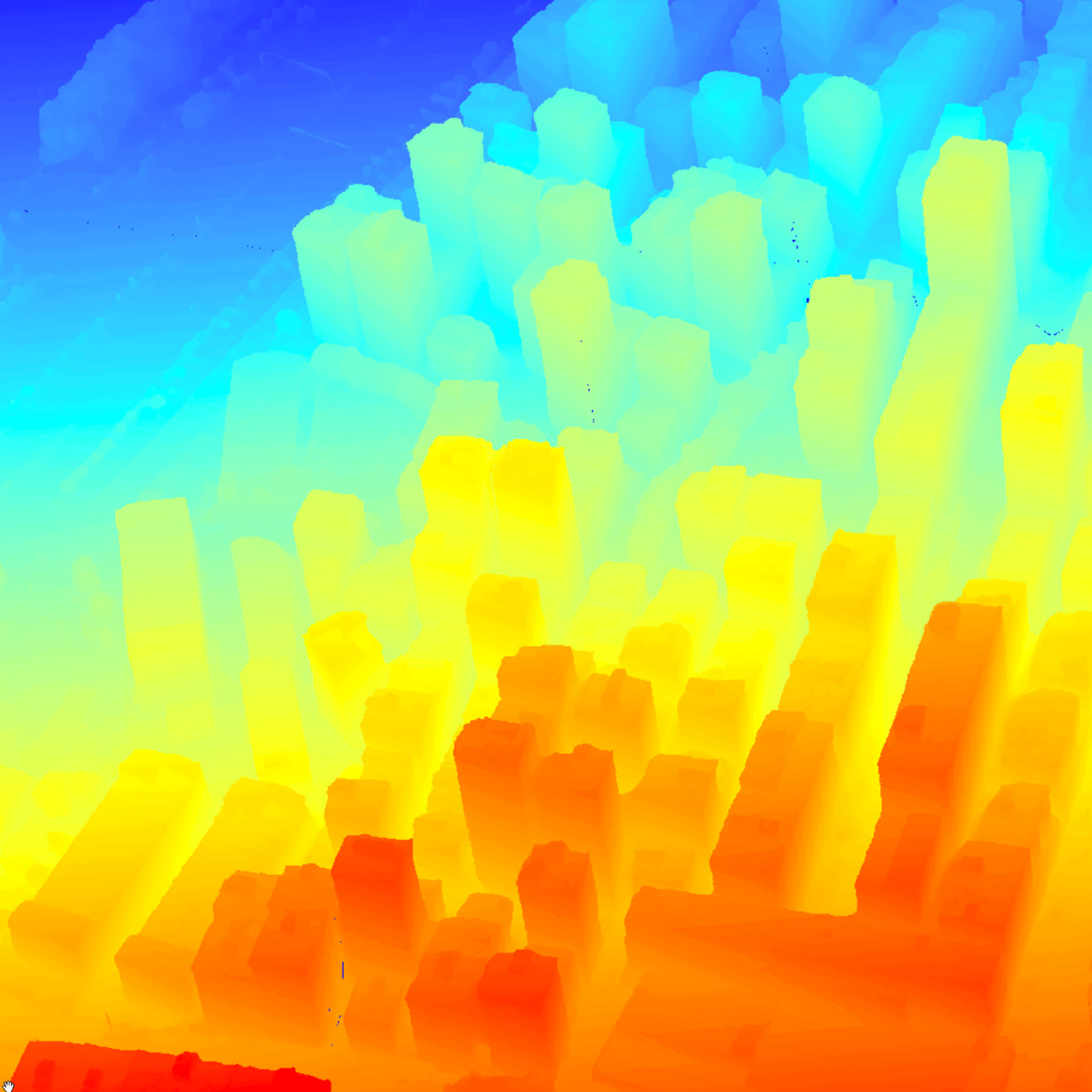}
    }
    \subcaptionbox{\textcolor{red}{Size} differences}[0.3\linewidth]{
        \includegraphics[width=\linewidth]{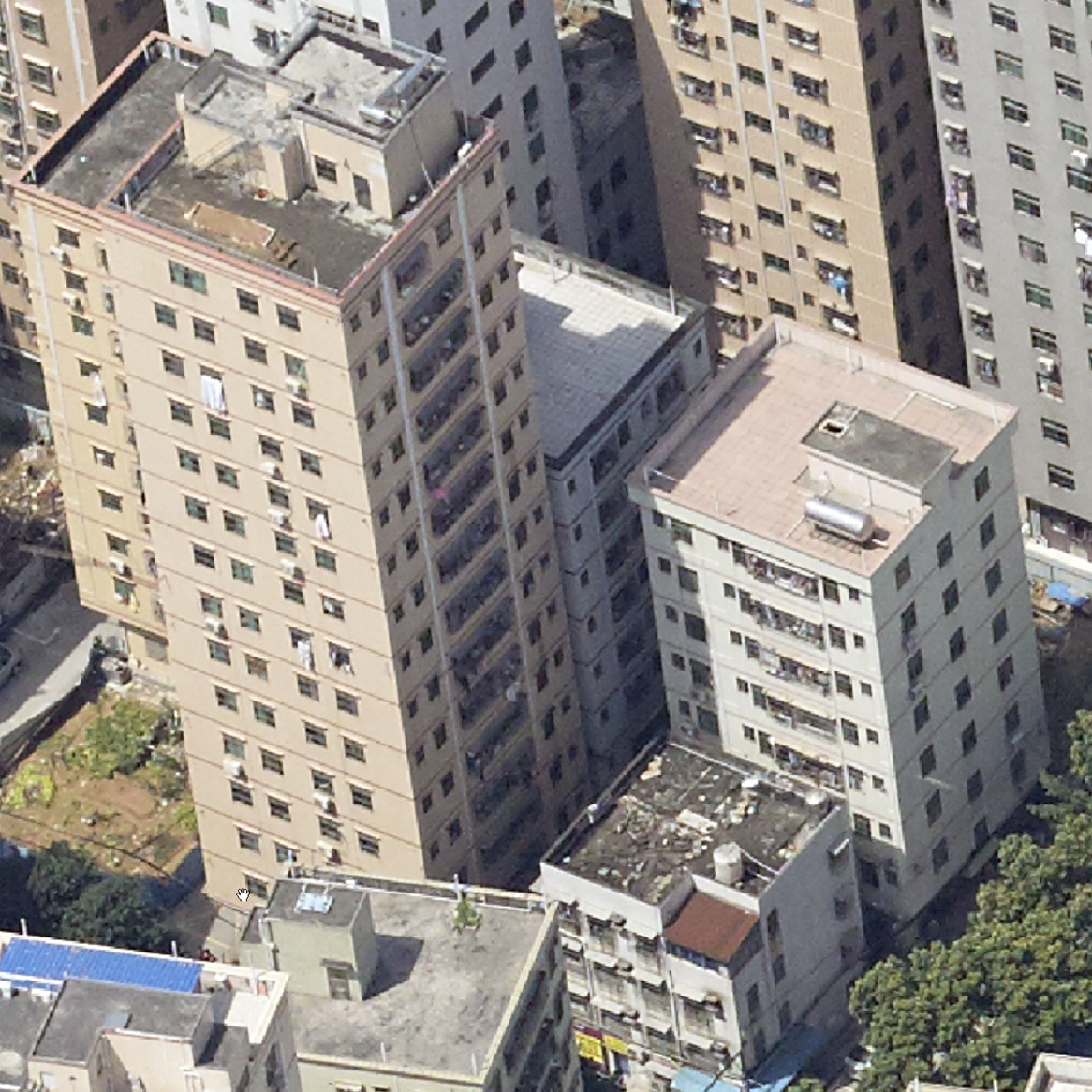}\vspace{0.3em}
        \includegraphics[width=\linewidth]{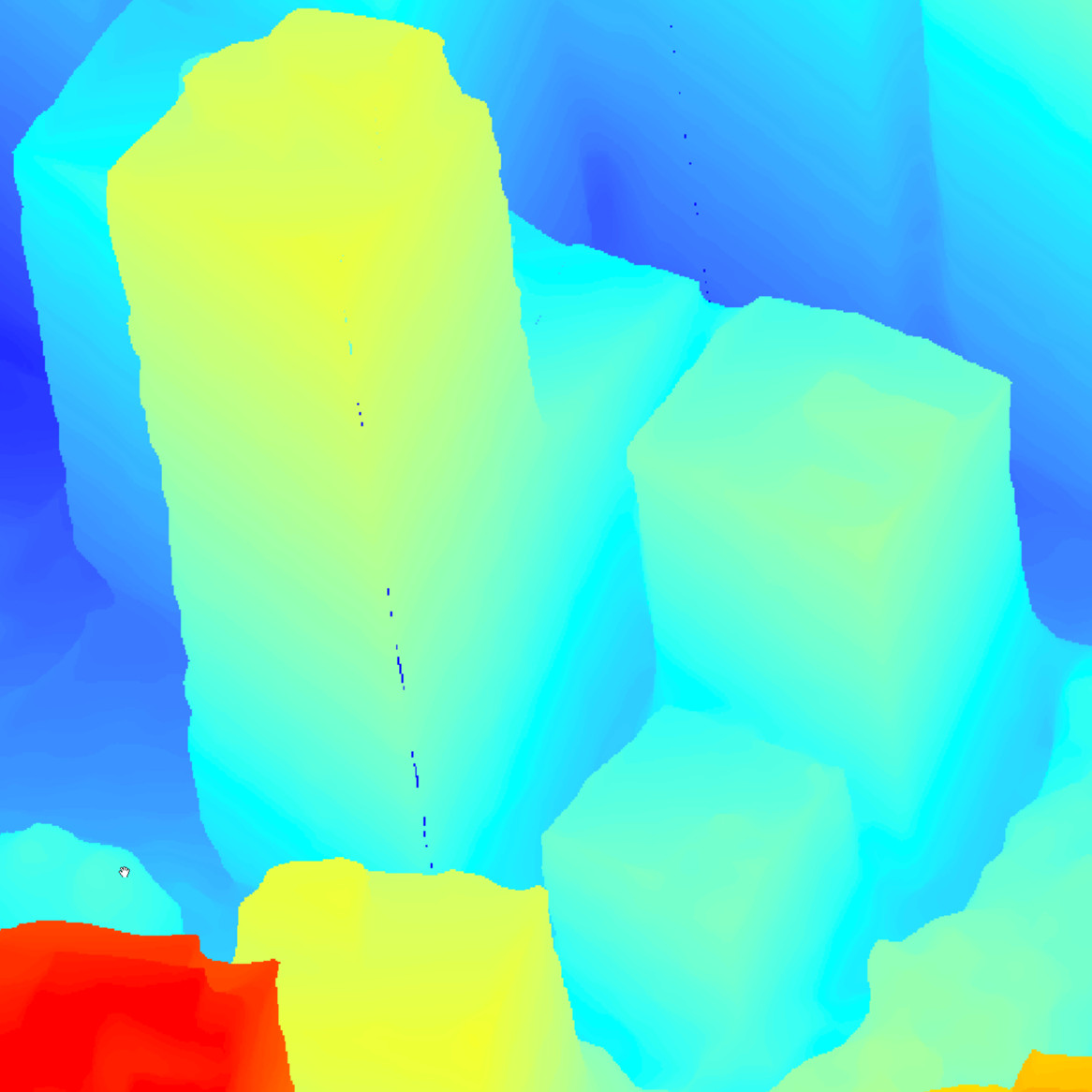}
    }
    \subcaptionbox{Occlusions}[0.3\linewidth]{
        \includegraphics[width=\linewidth]{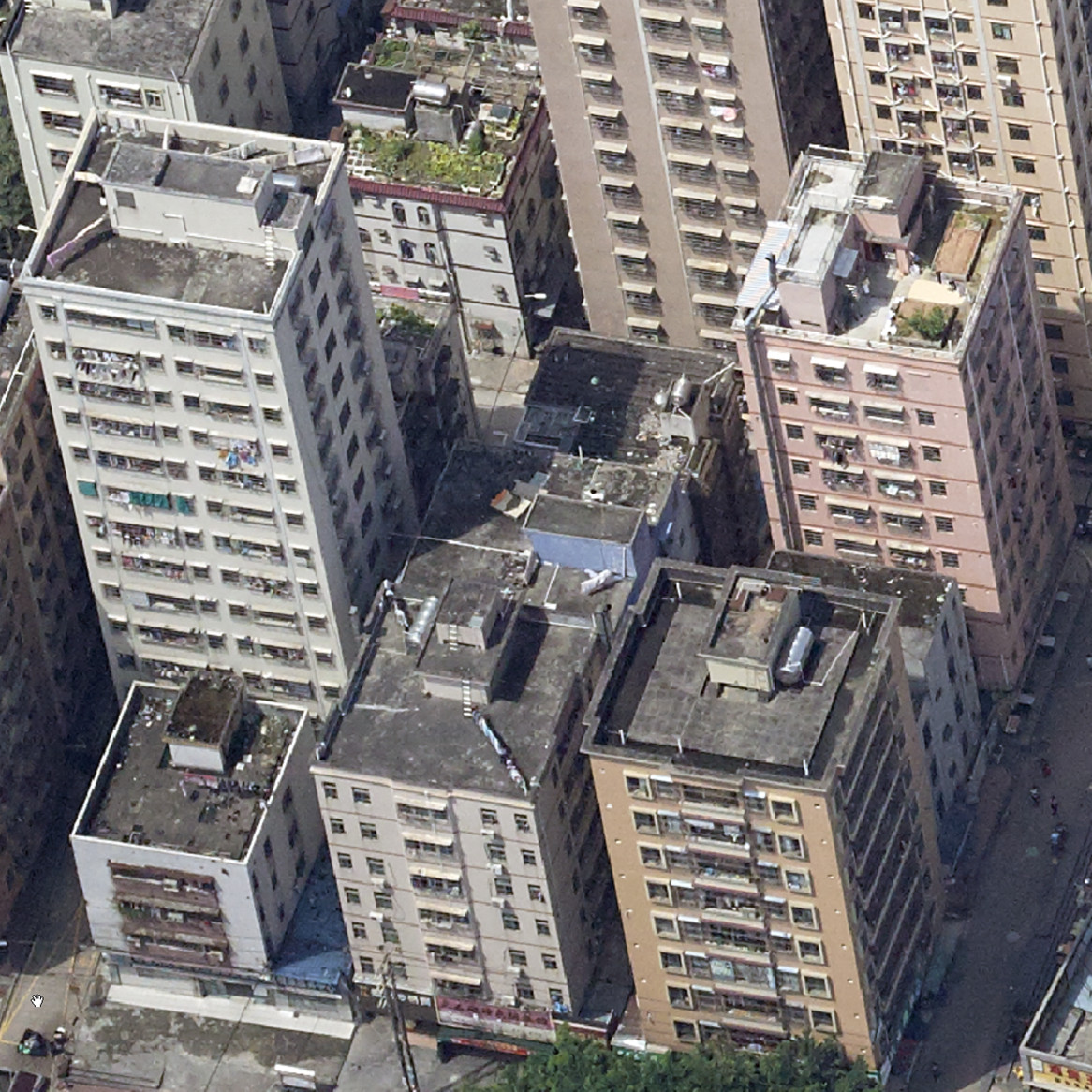}\vspace{0.3em}
        \includegraphics[width=\linewidth]{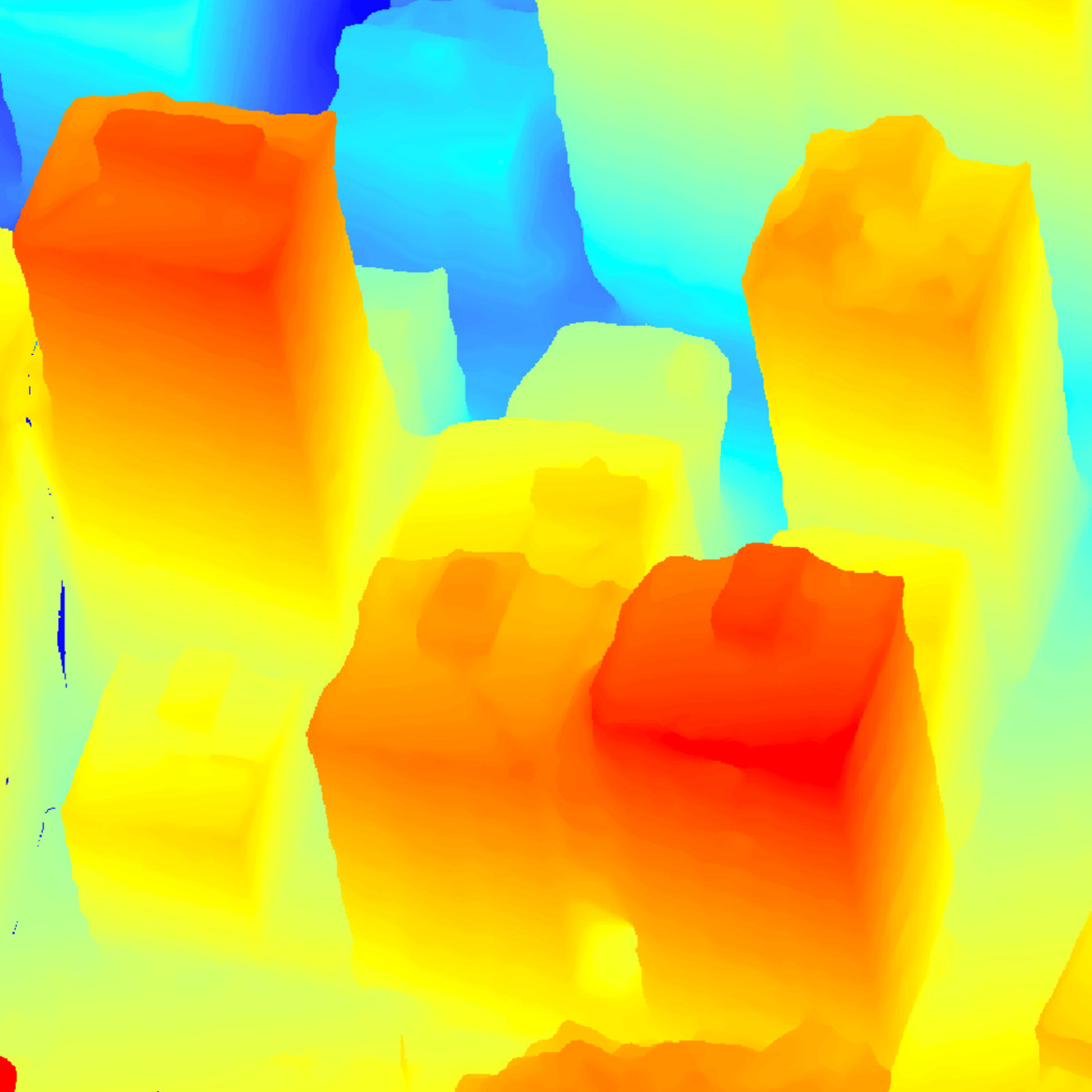}
    }
    \caption{Problems of scale differences and occlusions for fa\c{c}ades. The first and second rows represent the color images and corresponding depth information, respectively. It should be noted that depth information can also be directly used for pixel-wise occlusion tests.}
    \label{fig:problem_obl}
\end{figure}

\paragraph{1) Discriminability of features from 2D images}
\red{Because a single 2D image does not contain depth information,} the performance of either meticulously engineered features \citep{xiao2015building,konstantinidis2016building} or learned features \citep{chen2017learning,zhang2017building} significantly decreases in regions with ambiguous textures (Figure \ref{fig:problem_nadir}).
Even when oblique images are considered, most approaches only use the oblique view as 2D images \citep{xiao2015building}, which fails to alleviate the problem (Figure \ref{fig:problem_obl}).
It has been proved that \red{accurate depth information, such as digital surface model (DSM), can help identify buildings, even when only nadir views are used \citep{rottensteiner2007building,qin2014change,zhou2020lidar}.}
Inspired by this strategy, we go one step further in exploiting depth information for oblique-only views by using photogrammetric mesh models \citep{remondino2013dense}, as shown in the second rows of figures \ref{fig:problem_nadir} and \ref{fig:problem_obl}.
We also use depth maps to detect occlusions in complex urban environments.

\paragraph{2) Size variation of buildings}
Buildings in complex urban areas have different \red{sizes}.
Moreover, oblique camera images show notable scale differences between the near and far sides, owing to the large pitch and roll angles.
Even for the same building, the scale will be different on images captured from different angles.
The generalization ability of existing engineered features \citep{nyaruhuma2012verification_2d,yang2015building} is doubtful for images at different scales and different noise-level.
Furthermore, currently different strategies are generally used for roofs and fa\c{c}ades \citep{xiao2015building}.
A unified strategy is preferred to handle both primitives.

\vspace{1em}

To address the \red{two} aforementioned problems, we propose a depth-enhanced feature pyramid network (FPN) for the occlusion-aware verification of buildings from oblique images.
The basic approach is to first enrich the oblique aerial images with depth information from photogrammetric meshes \citep{zhu2020leveraging}; then, use depth-enhanced penta-view images for building verification.
More specifically, 3D mesh models are first reconstructed using automatic structures from motion and multi-view stereo pipelines \citep{schoenberger2016sfm,vu2011high}.
Next, depth information for all the images is retrieved by rendering the mesh models into the corresponding views \citep{zhu2020leveraging} using known camera poses.
Next, the outlines of buildings in the existing database are extruded according to associated attributes (e.g., number of floors) and projected to all visible views; then, visible patches are extracted through direct depth-information-based occlusion tests.
We establish a depth-enhanced FPN \citep{lin2017feature} with an additional fusing layer, to classify the visible patches (which have various scales) to three categories: roof, fa\c{c}ade, and background.
Finally, a robust multi-view voting strategy is used to ensure that all changed buildings are marked, with a relatively small false-alarm rate set as an ancillary goal. 

\red{To summarize, in order to alleviate problem caused by false alarms, the primary contributions of this paper are : 1) An efficient approach to associate 3D mesh models with 2D images, which makes the features more discriminative and supports occlusion tests.
2) An improved pyramid scheme \citep{lin2017feature} tuned for oblique images for buildings of various \red{sizes}.
3) A robust multi-view voting strategy, which achieves a 100\% true-positive rate with a relatively small false-alarm ratio in two datasets containing more than 3000 buildings.}
The reminder of this paper is organized as follows:
Section \ref{s:related} briefly reviews the existing studies on building verification and the related works on convolutional neural networks (CNNs).
Section \ref{s:methods} describes the depth-enhancement strategy and multi-view verification procedure in detail. 
Section \ref{s:experiments} describes the experimental evaluations, and Section \ref{s:conclusion} presents our concluding remarks.

\section{Related works}
\label{s:related}

In the following, we briefly review the most pertinent topics of this paper: 1) building verification, 2) image classification with learned features, and 3) multi-scale approaches.

\paragraph{1) Building verification}
Earlier works \citep{rottensteiner2007building,singh2012building} on building verification typically used the normalized difference vegetation index to perform threshold screening of roof areas.
However, such methods are susceptible to interference from vegetation-covered areas or roofs.
Several solutions have been proposed to improve the method's robustness, including principal component analysis \citep{deng2008pca}, texture features \citep{sidike2016automatic,sofina2016building}, and shape analysis  \citep{abdessetar2017buildings}.
However, these methods cannot overcome the ambiguities caused by ground-object interference.

The re-invention of oblique aerial imaging \citep{remondino2015oblique} has made it feasible to consider fa\c{c}ades as an additional source of information for improved building verification.
Consequently, doors \citep{nyaruhuma2010evidence} and other building features \citep{xiao2012building} have been used for verification, based on geometrical inferences or texture information.
For example, lines or corners---the most salient features of building fa\c{c}ades---are typically used \citep{nyaruhuma2010evidence}.
Texture information \citep{frommholz2015extracting,yang2015building,zhu2020interactive,hu2016texture} can also be used to parse fa\c{c}ades, and roofs and fa\c{c}ades can be jointly considered to improve robustness \citep{xiao2015building}.

However, most of these approaches use solely the oblique aerial images as ``pretty images'' and they neglect their inherent value in 3D measurement.
In fact, it has been proved that even using 2.5D depth information from the DSM will increase the discriminability of roof features \citep{rottensteiner2007building,zhou2020lidar}.
We expand upon this strategy by enhancing all images with depth information, rather than using only the orthophoto \red{and the DSM}.

\paragraph{2) Image classification with learned features}
Most of the aforementioned building-verification methods use meticulously engineered building features (e.g., lines, corners, and textures), which may not generalize well to different scenarios \citep{huang2016building,hu2016stable,hu2017bound}. 
With the advent of deep learning approaches (especially deep CNNs \citep{krizhevsky2012imagenet}), learned features have shown impressive performances in numerous tasks \citep{simonyan2014very,he2016deep,alshehhi2017simultaneous,zhu2020unsupervised}.
Many approaches have been proposed to solve the related topic of change detection in buildings, based on the Siamese network\citep{bromley1994signature,bertinetto2016fully}.
For instance, \cite{zhan2017change} directly converted the Siamese network into a change-detection one, using aerial images as the input; \cite{zhang2018change} generalized a similar approach to manage multi-modal data sources.
Another strategy is to extract building areas using instance segmentation \citep{ji2019building} (e.g., Mask Regions with CNNs \citep{he2017mask}) and compare the binary masks using U-Net \citep{ronneberger2015u}.

\red{However, we argue that formulating the problem of building verification in terms of change detection is not ideal.
Unlike the orthophotos with nadir views, oblique views cannot be rectified globally and therefore it is hard to perfectly align them from two-phase images.
Only the roof structures can be considered for pixel-wise change detection.}
Therefore, we directly formulate the verification as an image-classification problem \citep{simonyan2014very,he2016deep}; this involves projecting the extruded building \red{planes} onto multi-view images and classifying the patches as roofs or fa\c{c}ades.

\paragraph{3) Multi-scale approaches}
Scale differences are a very common issue in many applications, and pyramid schemes have already been widely applied to resolve them.
For instance, spatial pyramid pooling has been proposed \citep{he2015spatial} to aggregate features across multiple scales.
A FPN \citep{lin2017feature}, constructed using up-sampling and lateral connections, has demonstrated an impressive performance in object detection.
The cascaded pyramid network formulation \citep{chen2018cascaded} also provides an efficient strategy for integrating features into pyramid networks.
In \cite{tan2019efficientnet}, pyramid scheme was learned using an automatic neural architecture search, which combined depth, width, and resolution transformations.
Furthermore, \cite{tan2020efficientdet} demonstrated a more efficient strategy for connecting between multi-scale features, to improve object-detection performances.
To solve the specific problems raised by conducting building verification using oblique aerial images, we adopt a strategy similar to the FPN \citep{lin2017feature}.
\red{Specifically, we propose a fusion layer to integrate multi-scale features for the subsequent classification of buildings.}

\section{Methodology}
\label{s:methods}
\subsection{Overview}

\begin{figure}[H]
    \centering
    \includegraphics[width=\linewidth]{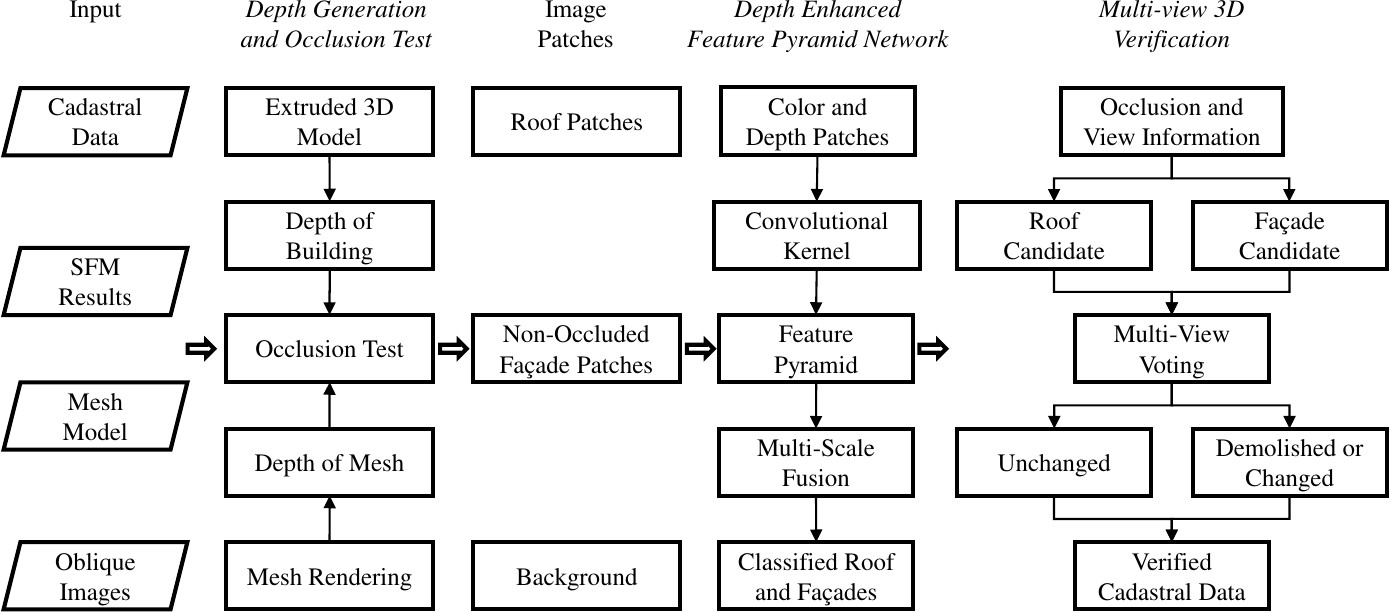}
    \caption{Workflow of the verification process for existing building data.}
    \label{fig:flowchart}
\end{figure}

To solve the problem of ambiguous textures using only orthophotos, we combine oblique images and depth features rendered from 3D mesh models for building verification.
We construct a FPN to extract features from the oblique and depth images; then, we fuse multi-scale features to improve the discriminability of buildings with varying sizes.
In addition, building verification reliability can be further improved by considering the redundancy of multi-view oblique images through a robust multi-view voting strategy.
The workflow of the proposed method is briefly illustrated in Figure \ref{fig:flowchart}, including the steps for depth generation and occlusion testing, the depth-enhanced FPN, and multi-view 3D verification.

\red{More specifically, this study first uses the mesh model generated from the multiple view stereo (MVS) \citep{hirschmuller2005accurate,vu2011high} pipeline and the position and orientation parameters from the Structure from Motion (SFM) \citep{schoenberger2016sfm,xie2016asymmetric} results, to render the corresponding depth information \citep{zhu2020leveraging}}.
The buildings' footprint data are also extruded using the corresponding floor information; \red{then, the depth image is obtained by projecting the photogrammetric mesh models onto corresponding view.}
Pixel-wise occlusion tests are conducted using depths obtained from both buildings and mesh models.
Second, a pyramid convolutional neural network is constructed, to extract feature information from the original color images and rendered depth images.
The non-occluded patches are classified into roof, fa\c{c}ade, or background categories.
Finally, the reliability is further improved through multi-view voting.
\red{A single building consists of a roof and several fa\c{c}ade planes, and each plane is also visible in multiple views, because of the increasing overlap ratio and the penta-view design \citep{petrie2009systematic} of aerial oblique camera system nowadays.}
A robust strategy is used to achieve a 100\% true-positive rate with a moderate false-alarm ratio.

\subsection{Depth generation and occlusion testing}
\label{s:depthandocclusion}

\begin{figure}[H]
    \centering
    \includegraphics[width=1\textwidth]{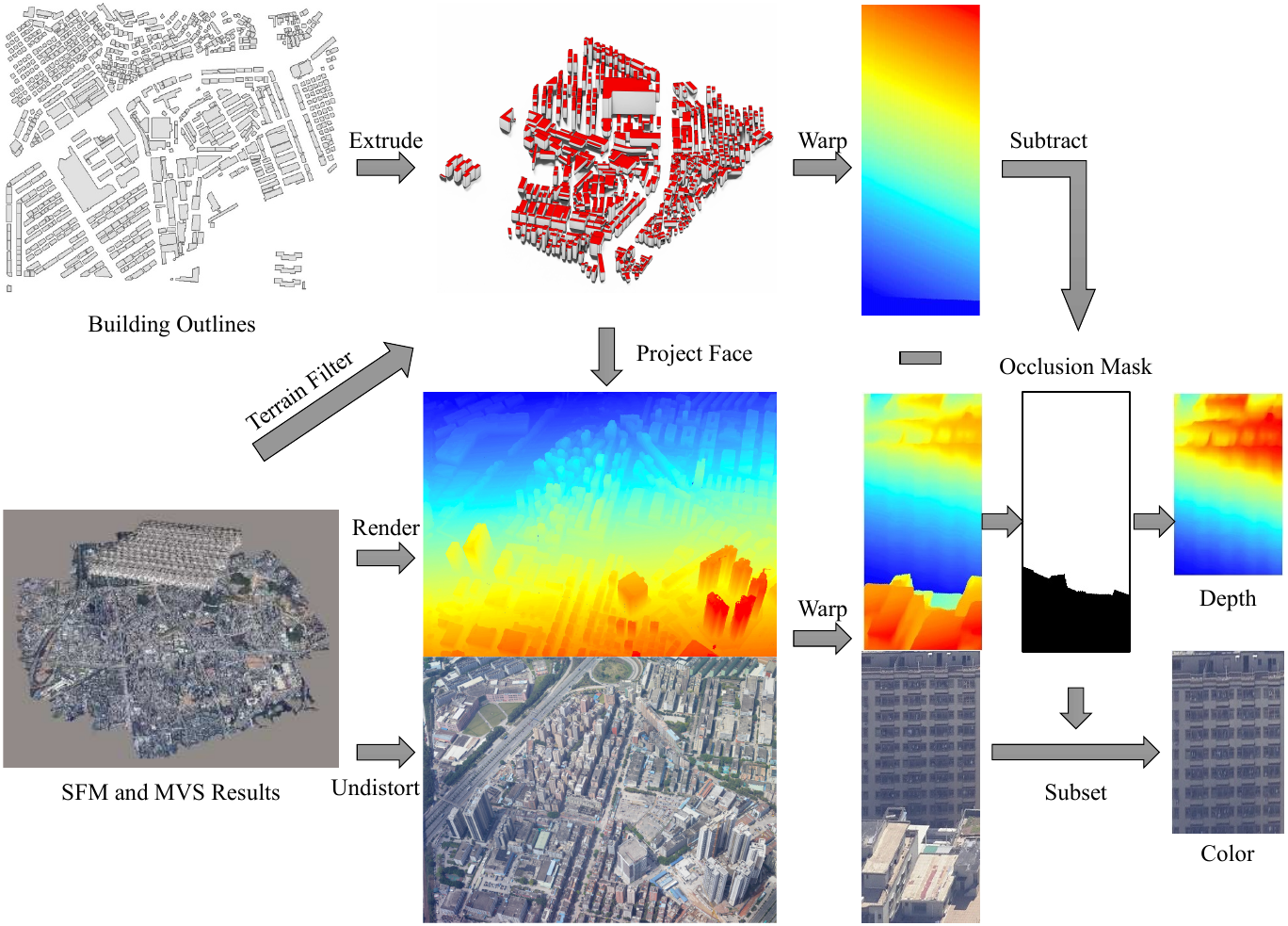}
    \caption{Generation of occlusion-free depth and color patches for roofs and fa\c{c}ades.}
    \label{fig:occlusion}
\end{figure}

Although the probabilities for multiple regions can be predicted using a unified approach \citep{girshick2015fast}, memory constraints limit the existing architecture to processing aerial images in only a single forward pass of the neural network.
In addition, occlusion is inevitable in building rise-up regions, especially in the fa\c{c}ade regions of oblique views.
Therefore, this study first subsets both roof and fa\c{c}ade patches for all buildings and then predicts the classes of each patch separately.
Furthermore, the depth information of the 3D mesh models is projected onto the corresponding aerial views.
Occlusion testing is conducted, and only visible patches are considered in the subsequent processing.
Figure \ref{fig:occlusion} illustrates the visible-patch-generation workflow.

\subsubsection{Generation of 3D buildings}
\label{s:gen3d}
Although many cities have migrated cadastral management data from 2D to 3D \citep{shojaei2013visualization}, 2D building floor plans \citep{fan2014quality} are still widely used in many applications for legacy reasons.
To use fa\c{c}ades for the verification of existing buildings, the 2D building outlines must be extruded to the corresponding level of detail (LOD-1) \citep{fan2014quality} representation.

The average heights of the mesh models or corresponding DSMs inside each building are used as the height of the roof.
To determine the height of the buildings, we first filter non-ground objects in the mesh model or DSM, using an adaptive surface filter \citep{hu2014adaptive}.
\red{An aggressive set of parameters is chosen for in this study, e.g., $200 \ m$ for maximum objects, $10 \ m$ for the minimum filtering window, and $0.2 \ m$ for the height threshold.}
After filtering out the ground, the roof is extruded to the lowest point inside the region, and a 3D LOD-1 model is obtained.
\red{An example of LOD-1 buildings generated from building outlines is shown in the top row of Figure \ref{fig:occlusion}.}

\subsubsection{Generation of depth and color images}
\label{s:gendepandcolor}

Because existing neural network architectures for 3D data (e.g., mesh \citep{hanocka2019meshcnn} and point clouds \citep{qi2017pointnet++}) are not scalable to city scale datasets, we propose an efficient approach to implement the 3D metric capabilities of oblique aerial images, by directly rendering the mesh models \citep{zhu2020leveraging} as pixel-wise depth maps.
This strategy is capable of handling tiled models \citep{osfield2004open} that are fragmented and discontinuous.

Similar to our previous work \citep{zhu2020leveraging}, we first convert the position and orientation information for each view (recorded in the BlockExchange format \citep{context2019camera}) to the corresponding notations in OpenGL \citep{glm2019opengl} (e.g., view $\mathbf{V}\in\mathbb{R}^{4\times4}$ and projection $\mathbf{P}\in\mathbb{R}^{4\times4}$ matrices).
A point $\boldsymbol{X}\in\mathbb{R}^3$ can be projected onto the normalized screen space $\boldsymbol{m}\in\mathbb{R}^3$ as 
\begin{equation}
    \tilde{\boldsymbol{m}}=\mathbf{P}\mathbf{V}\tilde{\boldsymbol{X}},
    \label{eq:oglproj}
\end{equation}
where the tilde symbols $\tilde{\boldsymbol{m}}\in\mathbb{R}^4$ and $\tilde{\boldsymbol{X}}\in\mathbb{R}^4$ denote the homogeneous coordinates.

We allocate a buffer to retrieve depth information (i.e., the third dimension of the normalized points $\boldsymbol{m}_z\in[-1,1]$ in the screen space), as shown in the center of Figure \ref{fig:occlusion}.
Because this rendering neglects the distortion parameters of the camera, the \red{aerial color image} is undistorted using the Brown distortion model \citep{context2019camera}.
In theory, the rendered depth and undistorted color images should possess a one-to-one mapping.
Therefore, we treat them as four channel inputs in the subsequent neural network.

\subsubsection{Occlusion testing and generation of depth and color patches}
\label{s:occlusionandgenpatch}

\begin{figure}[H]
    \centering
    \includegraphics[width=\linewidth]{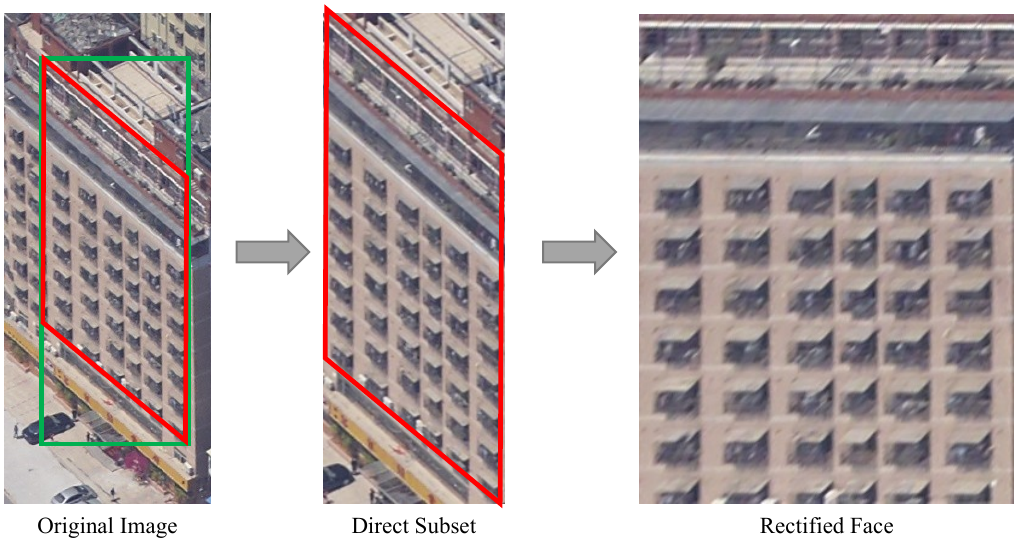}
    \caption{Rectifying the patches. The green rectangle marks the bounding box of the projected face on the original image, and the red polygon marks the actual region of the face after rectification.}
    \label{fig:warp}
\end{figure}

\red{In this section, we describe the procedure of occlusion detection and generation of image patches.
The vertices of each polygonal face of the extruded LOD-1 model are projected onto the undistorted image using Eq. \ref{eq:oglproj}.}
Instead of directly sub-setting the patches from the original images, we warp them to the polygonal face to alleviate affine deformations, as shown in Figure \ref{fig:warp}.
This warping procedure is defined by a homographic transformation $\mathbf{H}$ between the projected points on the image and the planar coordinates of the vertices \red{on the rectified face (Figure \ref{fig:warp})}, scaled by the average ground sample distance.

In the rectified image space, a pixel-wise depth map for the polygonal face is obtained through a bi-linear interpolation of the four corners.
Furthermore, the corresponding depths for the mesh model and original color image are also obtained using the homographic matrix $\mathbf{H}$, as shown in the right-hand side of Figure \ref{fig:occlusion}.
It should be noted that the bottom of the target building is often occluded by buildings in the foreground area.
Using both the depth map of the polygonal face and the mesh models, the occlusion test can be readily conducted by directly subtracting the two maps.
Then a binary mask is generated by thresholding the difference, \red{two meters} is used empirically.
Only the axis-aligned bounding boxes of the non-occluded subsets are used in the subsequent neural network, as shown in the right-most depth and color patches in Figure \ref{fig:occlusion}. 

\subsection{Fused FPN for building recognition}
\label{s:ffpforbuilding}

\red{Existing CNN architectures, e.g., VGGNet \citep{simonyan2014very} and residual neural network (ResNet) \citep{he2016deep}, have already demonstrated impressive performances on large datasets \citep{krizhevsky2012imagenet}.}
However, we found that after fine tuning, the performance of these models remained unsuitable for our task: to identify all changed buildings with a \red{relatively low} false-alarm rate.
When this objective is achieved, quality control is only required for buildings marked as changed by the method.

\begin{figure}[h]
    \centering
    \includegraphics[width=\linewidth]{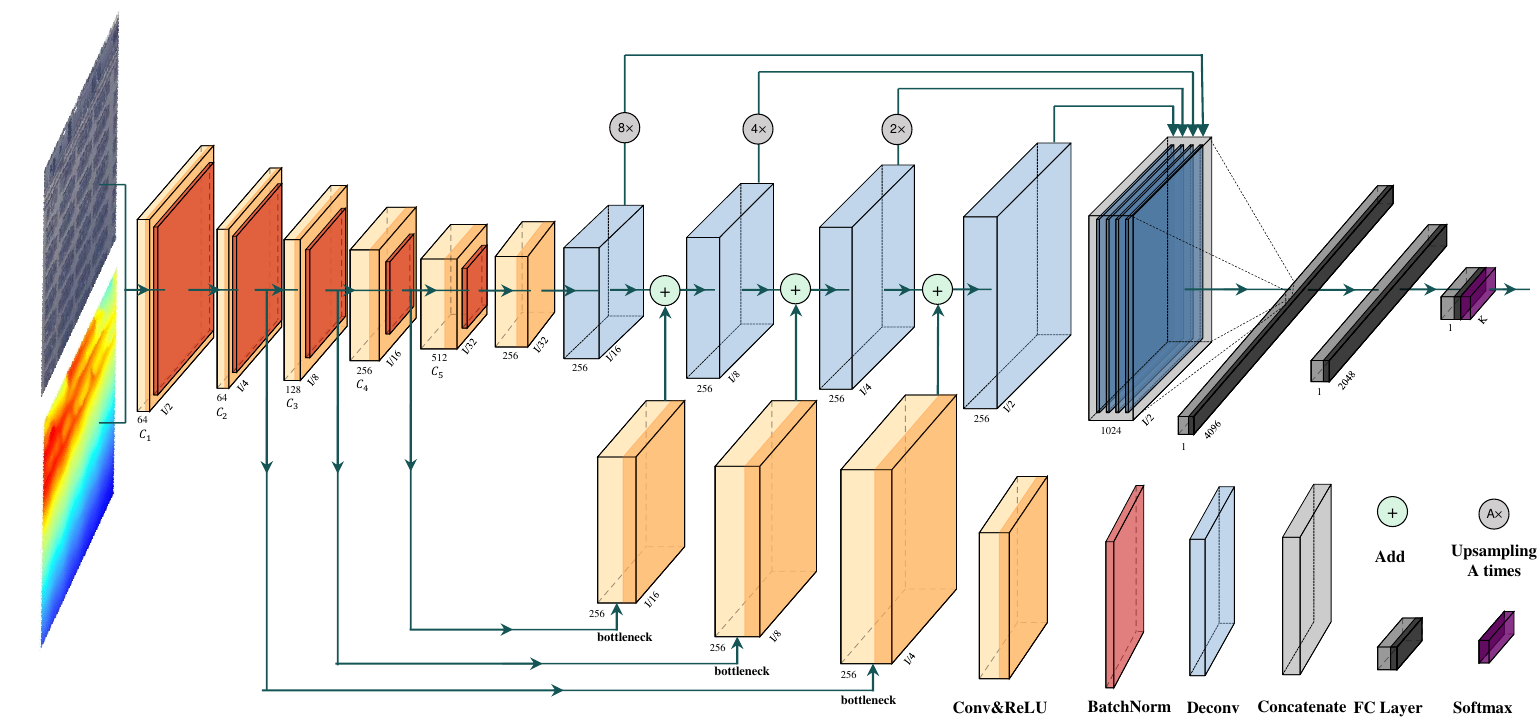}
    \caption{Depth-enhanced fused FPN architecture for roof and fa\c{c}de recognition.}
    \label{fig:network}
\end{figure}

To solve the problems caused by varying building sizes and scale changes in oblique aerial images, we propose a fused FPN, to learn a hybrid representation from both the color and depth images; this method is inspired by the FPN in \citep{lin2017feature}.
The purpose of constructing the pyramid is to obtain multi-scale features by decomposing the features of the image into large scales and small scales using the multi-scale convolution kernel of the convolutional neural network.
As shown in Figure \ref{fig:network}, we add a fused module to the tail of the architecture, rather than empirically choosing a suitable pyramid level in the FPN \citep{lin2017feature}.
In addition, the input is augmented by the corresponding depth image.
More specifically, we apply the corresponding numbers of $1 \times 1$ convolutional kernels, to reduce the dimensions of the small-scale feature maps; then, we use the deconvolution kernels \citep{long2015fully} to expand the sizes of the feature maps; finally, we fuse each of these using a large-scale convolution kernel, to generate the predictive feature map.
\red{The network ends with a global average pooling layer and a fully connected layer.
In addition, the \textit{softmax} activation is used to generate the normalized probability for each class}.
The network uses a cross-entropy loss function for multi-class recognition.

\subsubsection{Generation of feature pyramid}
\label{s:featurepyramid}

As shown in Figure \ref{fig:network}, we use a standard approach \citep{simonyan2014very,he2016deep} to generate a series of feature maps $\{C_1,C_2,C_3,C_4,C_5\}$ as the neural network input.
The number of channels and spatial resolutions of the feature maps gradually increase and decrease as $\{64,64,128,256,512\}$ and $\{I/2,I/4,I/8,I/16,I/32\}$, respectively; $I$ is the size of the image.
For each convolution block, Rectified Linear Unit (ReLU) activation and batch normalization \citep{ioffe2015batch} are used.

Because of the large spatial resolution of the feature map $C_1$, only the subsequent maps $\{C_2,C_3,C_4,C_5\}$ are chosen as the convolutional layers of the multi-scale feature pyramid structures, as shown in the left-hand side of Figure \ref{fig:network}.
\red{In addition, the feature map channels $\{C_2,C_3,C_4,C_5\}$ are resized to the same channels that is $256$ in the bottleneck,  using a $1\times 1$ convolution layer before fusing (the yellow feature map in the middle-bottom of Figure \ref{fig:network}), which is also called lateral connections in FPN \citep{lin2017feature}.
To retain the strong semantic information in the small-scale feature map, we use a deconvolution structure (the blue feature map in the center of Figure \ref{fig:network}) to scale up each pyramid layer from high to low.
Then, we add identically sized feature maps together after the $1\times 1$ convolutional layer in the former procedure; subsequently, the feature maps of all pyramid sizes contain larger quantities of strong semantic information.
It should be noted that the structure is also unaffected by the choice of convolution block, e.g., VGG \citep{simonyan2014very} or ResNet \citep{he2016deep}.
In this study, ResNet is chosen empirically as the base module for FPN, as it is well modularized and has been widely used in the remote sensing community.}

\subsubsection{Fusion of feature pyramid}
\label{s:fusionpyramid}

Although the feature pyramid module is constructed from the learned features in each layer, the different sizes and shapes of buildings make it difficult to effectively portray the features in a single layer. 
Therefore, we use a large convolutional map as a baseline, to retain more building details in the convolutional map.
By integrating the features of the pyramids at each level, the network can achieve a balanced implementation of these features at each layer.
More specifically, we first expand the scales of the feature maps at each pyramid level via upsampling (the gray operator shown in the middle-top of the Figure \ref{fig:network}), to retain the feature maps structures as much as possible.
Then, we concatenate all layers, as shown in the right-hand side of Figure \ref{fig:network}.
The final concatenated feature map is fed into a fully connected layer and activated by a \textit{softmax} layer before being classified as roof, non-occluded fa\c{c}ade, or background.
A multi-class cross-entropy loss is used to train the network.

\subsubsection{Implementation details}
\label{s:implementation}

In this study, the training samples consisted of 6000 evenly sampled images for the roof, fa\c{c}ade, and background.
The sample images were resized to $224\times 224$, the input size of the network.
The dataset was randomly divided into training and testing sets in a $7:3$ ratio.
A pre-trained ResNet network \citep{he2016deep} from ImageNet was used.
For the depth layer, the average of the three color channels was used in the first layers convolution kernel, and the remaining parameters were zero-initialized.
A total of 50 epochs were set for training.
The initial learning rate was 0.1, with a five-fold decay every ten epochs.
The model with the optimum testing accuracy was used.

\subsection{Multi-view voting strategy for building verification}
\label{s:verification}

\begin{figure}[H]
    \centering
    \includegraphics[width=\linewidth]{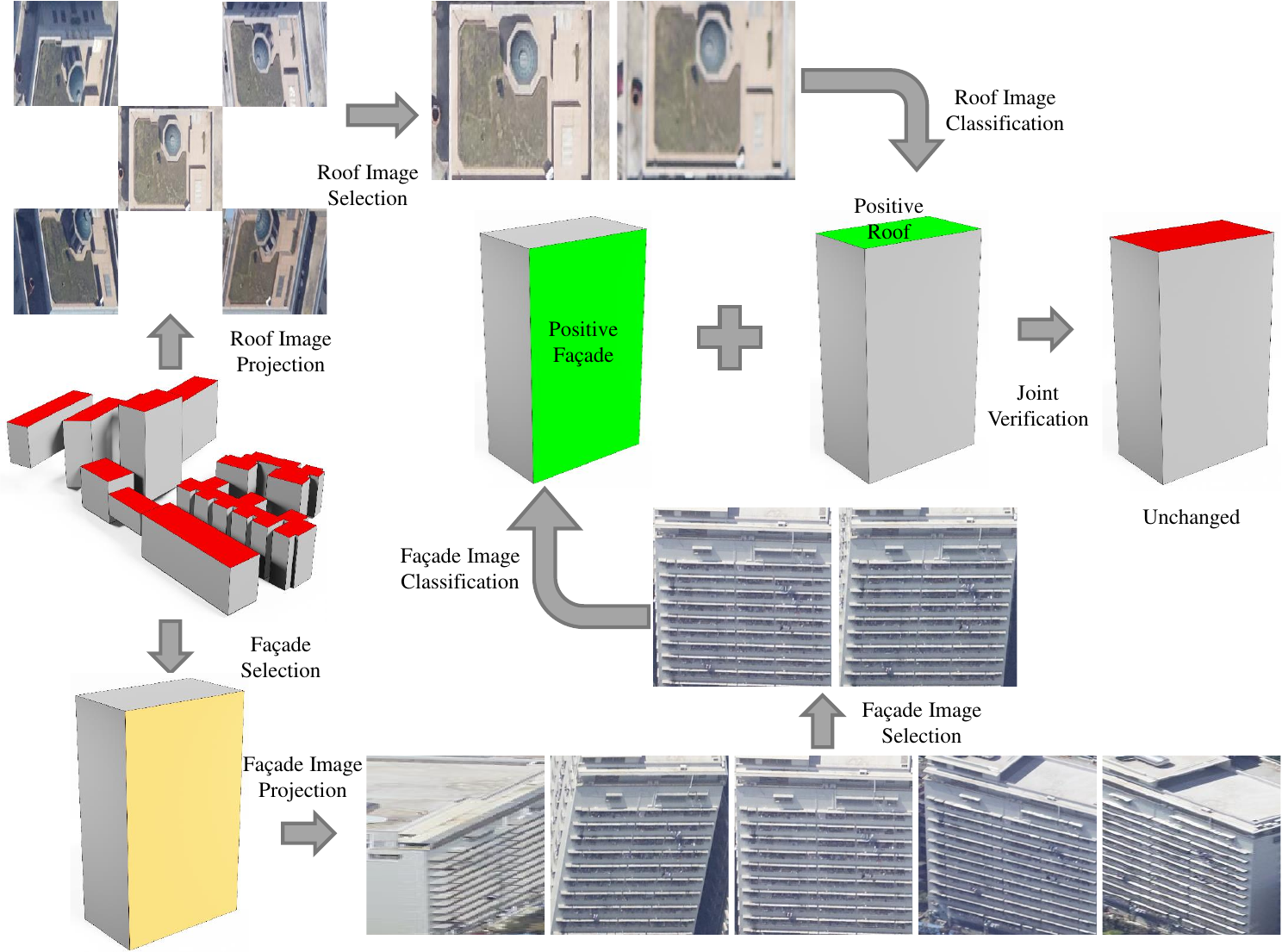}
    \caption{Multi-view voting strategy for building verification. The verification process includes roof and fa\c{c}ade verification, as shown in the upper and lower parts of the figure, respectively.}
    \label{fig:verification}
\end{figure}

To improve the reliability of the overall building verification procedure, we designed a multi-view voting strategy for roofs and fa\c{c}ades, using the multi-view visibility of buildings in oblique images.
In Section \ref{s:occlusionandgenpatch}, when obtaining building image patches, the heights and locations in the building footprint data were not sufficiently accurate, which can result in large deviations of the projection images.
Therefore, to make the building verification process more robust, we constructed a building verification image set, \red{by selecting images that are directly facing the building, e.g., the angle between the bore-sight of the camera and normal vector of the plane is small}.
Because the projection area from images \red{with smaller angle difference (Figure \ref{fig:verification})} is often the largest, we used image-projection-area ranking to filter out off-center photographed building images, thus constituted the building verification image set.

The visible projected areas of the roofs in the images were typically square in the nadir view (as shown in the upper left-hand side of Figure \ref{fig:verification})
Thus, the well-projected central area roof images which should be the first $1/4$ \red{(start from the center point by reducing the length and width by half, resulting in $1/4$ total count)} of the projected area, are selected to constitute the verification set of the roof.
\red{The two larger fa\c{c}ades of the same building were selected based on the non-occluded area; although more fa\c{c}ades views are absolutely possible, this ensures that the necessary building structure is selected and avoids serious occlusions of the fa\c{c}ades.}
The projected visible area of the fa\c{c}ade in the image was typically a sector area from the oblique view (as shown in the lower right-hand side of Figure \ref{fig:verification}).
Thus, the well-projected central area fa\c{c}ade images which should be the first $1/3$ \red{(start from the central view by reducing the angle to $1/3$, resulting in $1/3$ total count)} of the projected area, are selected to constitute the verification set of the fa\c{c}ades.
Then, the image candidate sets for the roof and fa\c{c}ade were classified; if all images from both candidate sets were correctly classified, the original building was judged to be unchanged.

\red{The objective of this strategy is to ensure that $100\%$ of the structures verified as buildings are indeed so; that is, we want to ensure that all changed buildings can be detected. 
Inevitably, some unchanged buildings will be judged as non-building structures; that is, a certain false-alarm rate is tolerable for non-building structures.}

\section{Experimental evaluation and analysis}
\label{s:experiments}

The experimental evaluation and analysis is divided into five parts.
The first part presents a basic overview of the experimental dataset.
The second part presents and analyses the overall experimental results.
In the third part, we select some typical ambiguous regions to evalute the specific utility of our three main contributions.
Then, we compare the CNN network with different structures against the proposed fused feature pyramid (FFP) network with superimposed color and depth images.
The final section discusses the applicability and limitations of the method.

\subsection{Datasets specifications}

\begin{figure}[H]
    \centering
    \includegraphics[width=1\textwidth]{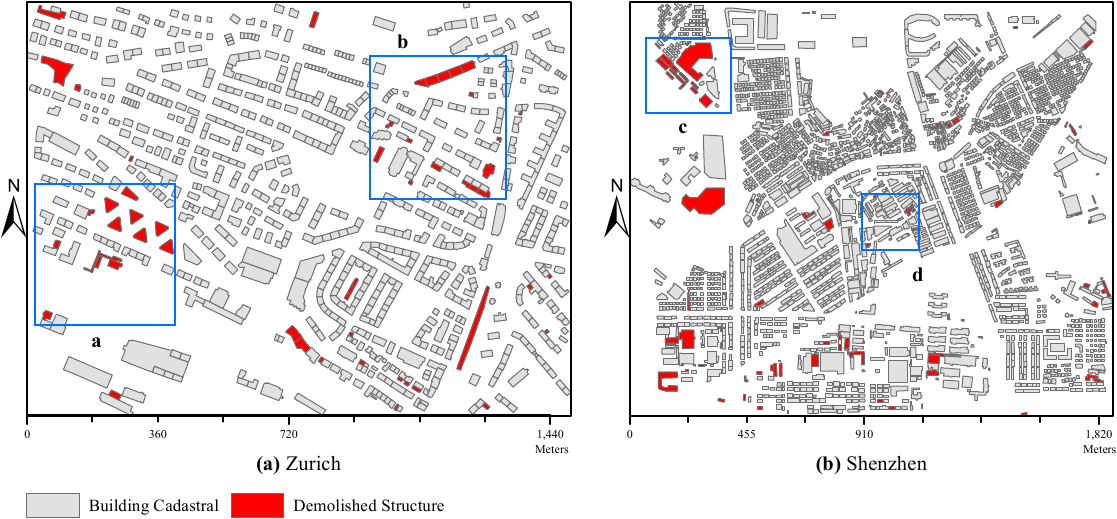}
    \caption{{Building footprint data} of oblique images in Zurich and Shenzhen. The gray mask denotes unchanged {building footprint data}, and the red mask denotes demolished structures. {Note the different scales of the two samples. The blue squares mark sub-samples described later on.}}
    \label{fig:experimental_area}
\end{figure}

In this paper, two datasets are used for experimental verification:
oblique image data of Zurich, collected in 2014 and compiled in the International Society for Photogrammetry and Remote Sensing Benchmark dataset\citep{cavegn2014benchmarking}; 
and oblique image data of Bantian District, Shenzhen, collected in 2016 by the Shenzhen Research Center of Digital City Engineering (SRCDCE).
The experimental area of the Zurich dataset was approximately $0.74 \ km^2$ and contained a total of 817 buildings.
\red{The building footprint data}, downloaded from OpenStreetMap \citep{OpenStreetMap}, were used to verify the algorithms performance; these data include 54 wrongly labeled or demolished building structures, as shown in Figure \ref{fig:experimental_area}(a).
The experimental area of the Bantian District dataset was approximately $2.52 \ km^2$ and contained a total of 2,506 buildings.
The census cadastral building data were collected in 2014; according to the SRCDCE, 114 structures had been demolished by 2016, as shown in Figure \ref{fig:experimental_area}(b).
More informations about oblique image dataset is listed in Table \ref{tab:dataset_description}.

\begin{table}[H]
	\centering
	\caption{\textcolor{red}{Overview of the two datasets.}}
	\label{tab:dataset_description}
	\begin{tabular*}{\linewidth}{c@{\extracolsep{\fill}}cc@{}}
		\toprule
		Dataset                       & Zurich        & Shenzhen        \\ \midrule
		Area ($km^2$)                 & 0.74          & 2.52            \\
		Buildings                     & 817           & 2506            \\
		Maximum density ($1/km^2$)    & 1104          & 1429            \\
		Building source               & OpenStreetMap & SRCDCE          \\
		Camera                        & Leica RCD30   & Phase One IQ180 \\
		Ground Sample Distance ($cm$) & 5-10          & 6-12            \\
		Overlap (Flight/Side)         & 70\%/50\%     & 80\%/60\%       \\
		Flight Direction              & East-West     & North-South     \\ \bottomrule
	\end{tabular*}
\end{table}

\red{We used ContextCapture \citep{acute3d} for aerial triangulation and mesh reconstruction assisted by ground control points provided by vendors of the datasets.
We empirically verified the quality of the aerial triangulation information, by checking the alignment of the 2.5-D depth and color images; for which, no obvious misalignment are detected.
We divided the whole regions into 59 and 143 grid tiles for Zurich and Shenzhen, respectively. Each tile requires approximately 2 hours to generate the mesh models.}
After projecting the \red{building footprint data}, corresponding training image datasets of oblique images were constructed for each experimental area.
The actual demolished building areas are marked in red in Figure \ref{fig:experimental_area}; these were checked manually as part of our evaluation methodology.

\subsection{Experimental results}

\begin{figure}[H]
    \centering
    \includegraphics[width=1\textwidth]{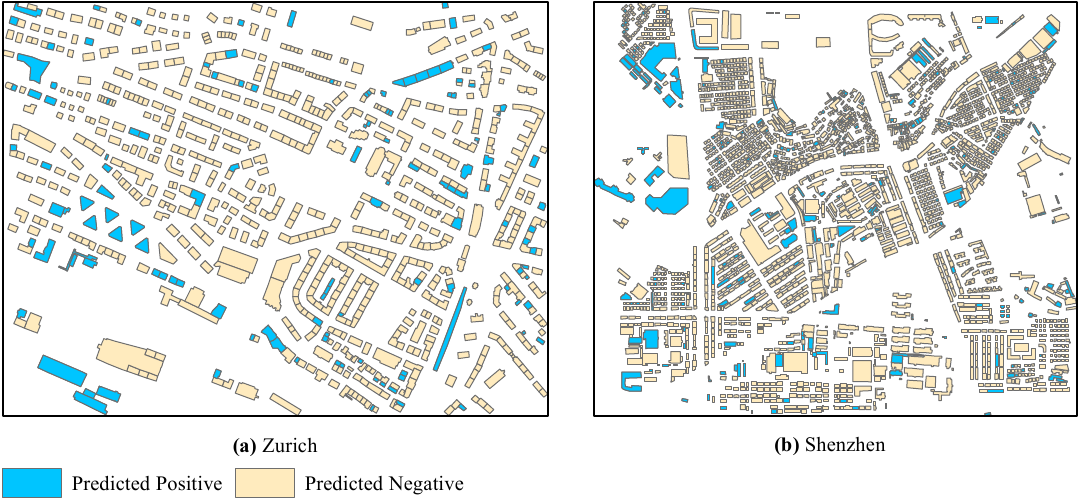}
    \caption{{Building footprint data} verified by multi-view voting strategy.}
    \label{fig:verify_dlg}
\end{figure}

The building types in Zurich are more regularly and sparsely arranged than those in Shenzhen;
this leads to differences in the quality of the oblique image projections, which in turn produces
more \red{variable} disturbance in image identification.
\red{All the hyper-parameters for the FPN network are kept the same in this work. 
To render the features of the building verification images more prominent, we need to select the proper value for the multi-view verification strategy and thereby select the optimal projection quality for the building image.
The numbers of selected fa\c{c}ade image (Figure \ref{fig:verification}) are selected differently for the two datasets, according to the visibility of fa\c{c}ade views.
Shenzhen is featuring dense, high and occluded buildings, therefore only the best fa\c{c}ade view is used; for Zurich, because the buildings are relatively lower and visibility better, two best views are used to improve robustness.}
An overview of the verified \red{building footprint data} obtained using the multi-view voting strategy for the Zurich and Shenzhen datasets is displayed in Figure \ref{fig:verify_dlg}.

Furthermore, to evaluate the performance of the proposed multi-view voting strategy, we compared the verification results from the nadir view (using only roof images) and oblique view (using only fa\c{c}ade images). 
We used the general statistical metrics of \red{TP (true positive), TN (true negative), FP (false positive) and FN (false negative)} as quantitative indicators.
Meanwhile, positive and negative samples refer to demolished and unchanged areas, respectively.
To evaluate the results more intuitively, we chose two ratios as criteria: the correctness of positive ($C_{P}$) and negative ($C_{N}$) results.
These are computed as:
\begin{equation}
    \begin{split}
        &C_{P} = \frac{TP}{TP+FN}\\
        &C_{N} = \frac{TN}{TN+FP}.
    \end{split}
\end{equation}
We \red{summarize} these metrics and display them in Table  \ref{tab:verification_result}.

\begin{table}[H]
    \centering
    \caption{{The correctness of positive and negative results for Zurich and Shenzhen datasets.}}
    \label{tab:verification_result}
    \begin{tabular}{@{}cccccccc@{}}\toprule
        Dataset&Viewing angle&TP&TN&FP&FN&$C_{P}$&$C_{N}$\\\midrule
        \multirow{3}{*}{Zurich}&Nadir view&51&717&46&3&94.4\%&94.0\%\\
                               &Oblique view&28&702&32&26&51.9\%&92.0\%\\
                               &\textbf{Multi-view}&\textbf{54}&\textbf{691}&\textbf{72}&\textbf{0}&\textbf{100\%}&\textbf{90.6\%}\\\midrule
        \multirow{3}{*}{Shenzhen}&Nadir view&62&2294&98&52&54.4\%&95.9\%\\
                                 &Oblique view&64&2323&69&50&56.1\%&97.0\%\\
                                 &\textbf{Multi-view}&\textbf{114}&\textbf{2231}&\textbf{161}&\textbf{0}&\textbf{100\%}&\textbf{93.2\%}\\\bottomrule
    \end{tabular}
\end{table}

The results indicate that in the Zurich experimental area, because of the uniformity of type and regular roof structures, most buildings could be verified from the nadir view (using only roof images).
When combined with the results of the oblique view verification, we could verify several demolished buildings that were only detectable from the oblique view. 
In the Shenzhen experimental area, the roof structures were complex, heavily covered by vegetation, and exhibited large differences in material; thus, it was difficult to comprehensively detect demolished buildings using roof inspection alone.
Meanwhile, this area featured extremely high building densities \red{(e.g., approximately 1430 buildings/$km^2$ in some places)} and considerable height differences between buildings; as a result, the fa\c{c}ade occlusion is so serious that some buildings are completely occluded.
In fa\c{c}ade detection, the optimal building fa\c{c}ade projection image should be selected for building verification.
When verifying building demolition, the results of the fa\c{c}ade verification can complement the results of the roof verification, whilst also improving the reliability of the overall verification.
However, it is inevitable that the relative rates of false detection for unchanged buildings increases with the integration of the roof and fa\c{c}ade verification results.
Therefore, it can be seen in Table \ref{tab:verification_result} that the correctness of the verification results for unchanged buildings in the \red{multi-view strategy} is \red{lower} than nadir view or oblique view.

\subsection{Experimental analyses}

In this section, we select some typical regions of ambiguity to evaluate the specific utility of our three main contributions.

\subsubsection{Effects of depth information}

\begin{figure}[H]
    \centering
    \includegraphics[width=1\textwidth]{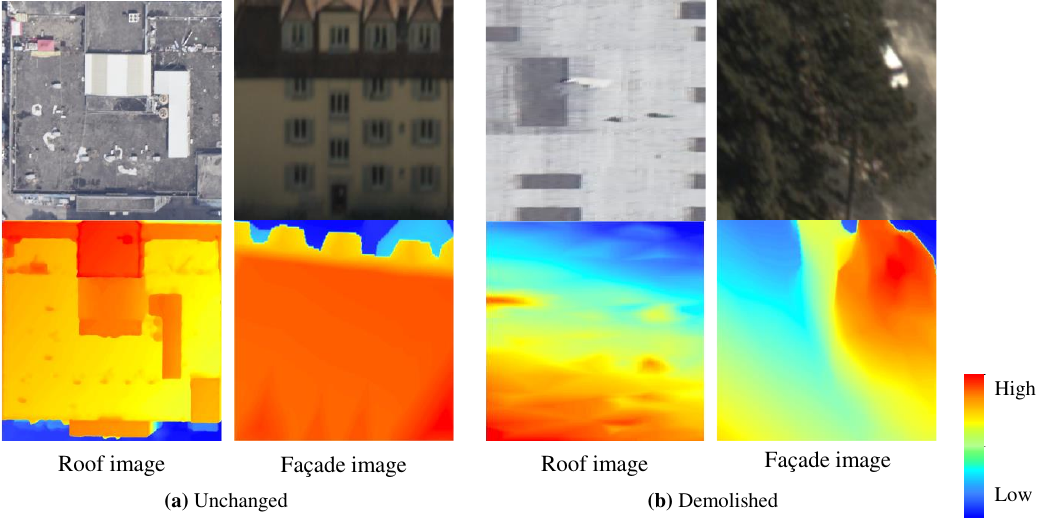}
    \caption{
        Color image and depth image comparison for demolished and remaining building. 
        In the color image, the features of the demolished building (b) are very similar to those of the unchanged building (a). 
        In the depth image, the features of the demolished building (b) differ from those of the unchanged building (a). {Note the ranges of depth for unchanged and demolished are different}. 
    }
    \label{fig:rgb_dep_compare}
\end{figure}

To interpret the characteristic differences between color images and depth images for the roof and fa\c{c}ade areas in buildings and non-buildings, Figure \ref{fig:rgb_dep_compare} depicts some typical patches.
The roof regions are fairly similar to the demolished remains seen from the nadir perspective, which makes it difficult to distinguish them using the traditional image feature descriptor. 
However, the depth image of the roof is more regular, whilst the distribution of depth information for non-building areas is more random.
Meanwhile, comparing the fa\c{c}ade images, we find that by performing occlusion detection of the depth information, we can filter out images with complete information for the fa\c{c}ade in the non-occlusion condition.
Therefore, we infer that depth information can provide a more effective basis for building classification.

\subsubsection{Effects of FFP module}

Analyzing the images of roofs in the Shenzhen area, we can see that they are well extracted and identified from large \red{sizes} ($3000 \ m^2$) down to small ones ($100 \ m^2$). 
We divided the building area \red{size} into three intervals: below $100 \ m^2$, between $100 \ m^2$ and $200 \ m^2$, and above $200 \ m^2$.
Dividing buildings in this way ensures that the number of buildings in each interval is evenly distributed.
\red{We counted the correctly verified building at each interval and calculated the correctness.
In the Zurich area, the correctnesses are 92.00\% for building size below $100 \ m^2$, 93.74\% between $100 \ m^2$ and $200 \ m^2$ and 95.65\% above $200 \ m^2$.
In the Shenzhen area, the correctnesses are 94.15\% for building size below $100 \ m^2$, 97.03\% between $100 \ m^2$ and $200 \ m^2$ and 95.96\% above $200 \ m^2$.
It could be noted that the correctness is almost not influenced by the sizes of buildings, thanks to the FFP module.}

\subsubsection{Effects of multi-view voting}

\begin{figure}[H]
    \centering
    \includegraphics[width=1\textwidth]{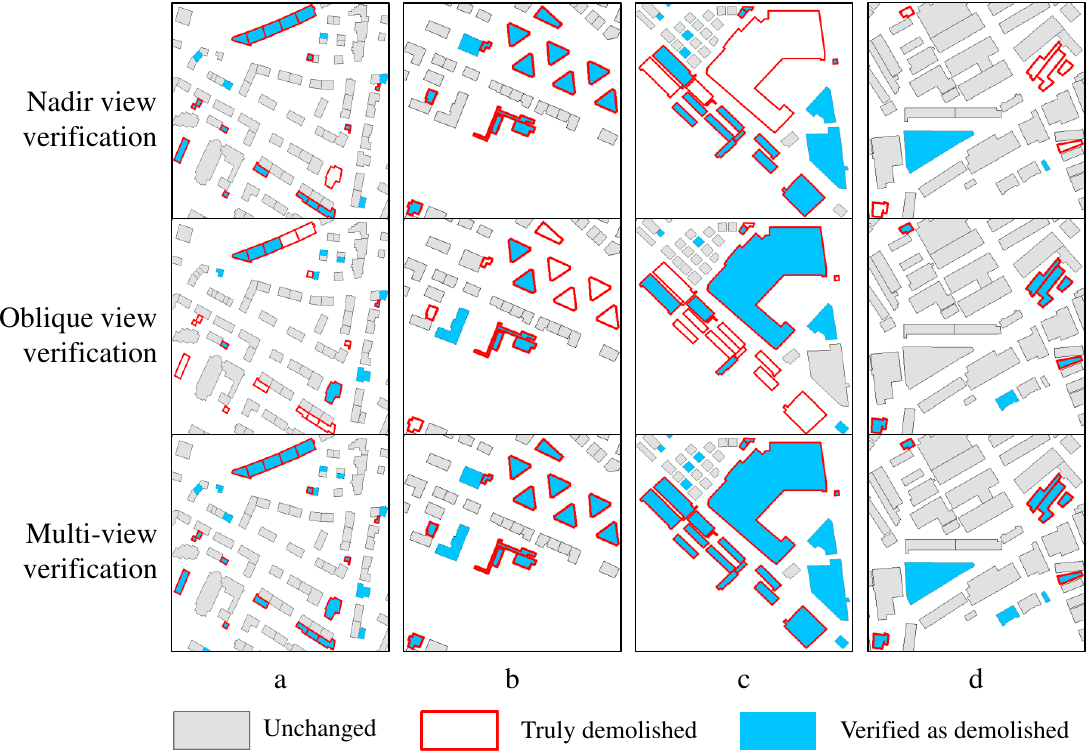}
    \caption{
        Nadir view, oblique view, and multi-view verification results comparison for demolished building.
        Several demolished areas could only be detected from the oblique view, owing to serious ambiguities in the nadir view {and vice versa}.
        The gray mask denotes the unchanged areas, the {red boxes} show the actually demolished buildings, and the blue marks denote buildings detected as demolished.
    }
    \label{fig:view_compare}
\end{figure}

We selected two typical areas from each of the two study regions, to illustrate the effects of multi-view voting; these are marked by the blue rectangles in Figure \ref{fig:experimental_area}. 
We compared the results verified using three different strategies: the nadir-view strategy (using only roof images), the oblique-view strategy (using only fa\c{c}ade images), and the multi-view strategy (using both roof and fa\c{c}ade images).
As shown in Figure \ref{fig:view_compare}, the large ambiguities observed from the single viewpoint meant that some demolished building could only be correctly determined using 3D verification, and using only nadir or oblique views tended to cause some kind of error.
We conclude that the \red{multi-view} verification strategy can achieve credible results.

\subsection{Ablation studies}
\label{s:ablation}

To further evaluate the effects of depth data and the FFP network, we conducted ablation studies using the two datasets.
First, the FFP module was compared against the well-established ResNet \citep{he2016deep} and the state-of-the-art EfficientNet \citep{tan2019efficientnet} models.
Second, for each architecture, we compared the network with and without the depth data.
From Table \ref{tab:ablation_experiment}, we conclude that the accuracy of each category in the FFP network is improved by more than $5\%$ compared to ResNet and more than $2\%$ compared to EfficientNet. 
Adding the depth convolutional channel leads to an accuracy improvement of more than $1\%$ for each type of network.
Therefore, the FFP network that integrates multi-scale feature pyramids with color and depth information is superior in classifying multi-size roofs and fa\c{c}ades.

\begin{table}[H]
	\centering
	\caption{{Ablation experiment results for each type of network. We use the the 34 layer ResNet and the B0 variant of EfficientNet for comparison. Beginning with the second row for each dataset, the values indicate the difference between the first row.}}
	\label{tab:ablation_experiment}
	\begin{tabular}{@{}clcccc@{}}
		\toprule
        \multirow{2}{*}{\textit{Dataset}} & \multirow{2}{*}{\textit{Network}} & \multicolumn{2}{c}{\textit{Roof}} & \multicolumn{2}{c}{\textit{Fa\c{c}ade}} \\ \cmidrule(l){3-6} 
		&                                   & Accu.(\%)       & Recall(\%)      & Accu.(\%)   & Recall(\%)   \\ \midrule
		\multirow{6}{*}{Zurich}           & FFP color+depth                   & 97.2            & 98.1            & 94.3        & 95.0         \\
		& FFP color                         & -0.9            & -0.5            & -1.1        & -0.2         \\
		& EfficientNet-B0 color+depth       & -0.4            & -1.5            & -1.6        & -2.2         \\
		& EfficientNet-B0 color             & -2.9            & -2.5            & -3.5        & -4.7         \\
		& ResNet-34 color+depth             & -5.0            & -5.7            & -5.0        & -5.5         \\
		& ResNet-34 color                   & -5.9            & -6.3            & -6.5        & -6.7         \\ \midrule
		\multirow{6}{*}{Shenzhen}         & FFP color+depth                   & 95.4            & 96.5            & 94.4        & 94.9         \\
		& FFP color                         & -2.2            & -3.1            & -1.0        & -1.9         \\
		& EfficientNet-B0 color+depth       & -0.5            & -0.4            & -1.7        & -3.1         \\
		& EfficientNet-B0 color             & -3.9            & -4.0            & -2.6        & -4.6         \\
		& ResNet-34 color+depth             & -5.0            & -6.5            & -4.4        & -4.7         \\
		& ResNet-34 color                   & -7.0            & -8.9            & -6.2        & -6.7         \\ \bottomrule
	\end{tabular}
\end{table}

\subsection{Discussion}

Based on the above evaluations of the FFP networks building verification abilities, we now discuss several characteristics and limitations of the proposed method.

\emph{1) Use of depth data.}
Existing oblique-image-based building verification methods do not take full advantage of the 3D information available in oblique images, e.g., the mesh models from the SFM \citep{schoenberger2016sfm,hu2016stable} and MVS \citep{vu2011high,hu2016texture} pipelines.
In this study, we fuse depth images with color images, to incorporate 3D depth information with 2D images in an occlusion-aware verification method.
The depth data eliminate the problems encountered in ambiguous texture regions.

\emph{2) Robustness of FFP and multi-view voting.}
The aim of the proposed method is to detect all changed buildings whilst maintaining a reasonably low false-alarm rate.
This requires the proposed method to be robust in different scenarios, and many redundancy checks are required.
This objective is achieved through the feature pyramid module and multi-view voting strategy, which uses the multi-view capabilities of oblique aerial images.

\emph{3) Limitations.}
As a supervised approach, manual work is still required for labeling roofs and fa\c{c}ades.
Furthermore, this method can only verify existing building data; it cannot detect new buildings from the oblique images.

\section{Conclusions}
\label{s:conclusion}

This paper proposes a multi-view voting strategy for oblique-image-based building verification using a depth-enhanced fused feature pyramid CNN.
We render the mesh model into the oblique image viewpoint, to obtain depth images.
Then, we combine the depth image and color image, to perform building verification using the fused FPN.
The fused multi-sized feature pyramid structure \red{improves} the accuracy of building-image classification.
The multi-view voting strategy also improves the building verification models performance.
\red{Future work may be devoted to building-change detection, not only for identifying demolished structures but also for discovering newly constructed ones.
To make the system robust, multi-view voting from oblique images and depth information from photogrammetric mesh models are probably still useful to improve the robustness.}

\section*{Acknowledgments}
\label{s:acknowledgments}
This work was supported in part by the National Natural Science Foundation of China (Projects No.: 41631174, 42071355, 41871291) and in part by the National Key Research and Development Program of China (Project No. 2018YFC0825803). 
In addition, the authors gratefully acknowledge the ISPRS/EuroSDR Project and the German Society for Photogrammetry, Remote Sensing, and Geoinformation for providing the Zurich dataset \citep{cavegn2014benchmarking}. 

\bibliographystyle{model2-names}
\bibliography{ffp}

\end{document}